\documentclass[letterpaper, 10 pt, conference]{icra/ieeeconf}  

\IEEEoverridecommandlockouts                              

\usepackage{amsmath} 
\usepackage{amssymb}  
\usepackage{amsmath,amssymb,amsfonts}
\usepackage{graphicx}
\usepackage{booktabs}
\usepackage{multirow}
\usepackage{pifont}
\usepackage{cite}
\usepackage{xcolor}
\usepackage[dvipsnames]{xcolor}
\usepackage{siunitx}
\usepackage{caption}
\usepackage{hyperref}
\newcommand{\cmark}{\ding{51}}
\newcommand{\xmark}{\ding{55}}

\title{\LARGE \bf
SHaRe-RL: Structured, Interactive Reinforcement Learning for Contact-Rich Industrial Assembly Tasks
}

\author{Jannick Stranghöner$^{1}$, Philipp Hartmann$^{1}$, Lisa-Marie Weigelt$^{1}$, \\ Marco Braun$^{2}$, Sebastian Wrede$^{1,2}$, Klaus Neumann$^{1,2}$}

\begin{document}
\raggedbottom
\bstctlcite{IEEEexample:BSTcontrol}

\maketitle
\thispagestyle{empty}
\pagestyle{empty}

\begin{abstract}
High-mix low-volume (HMLV) industrial assembly, common in small and medium-sized enterprises (SMEs), requires the same precision, safety, and reliability as high-volume automation while remaining flexible to product variation and environmental uncertainty. Current robotic systems struggle to meet these demands. Manual programming is brittle and costly to adapt, while learning-based methods suffer from poor sample efficiency and unsafe exploration in contact-rich tasks. To address this, we present SHaRe-RL, a reinforcement learning framework that leverages multiple sources of prior knowledge. By (i) structuring skills into manipulation primitives, (ii) incorporating human demonstrations and online corrections, and (iii) bounding interaction forces with per-axis compliance, SHaRe-RL enables efficient and safe online learning for long-horizon, contact-rich industrial assembly tasks. Experiments on the insertion of industrial Harting connector modules with 0.2–0.4\,mm clearance show reliable learning within practical wall-clock budget and improved performance over an unstructured human-in-the-loop RL baseline. We further show that the learned policy generalizes to previously unseen connector variants. Overall, our results show that process expertise alone can effectively guide real-world RL, making deployment safer, more robust, and economically viable for industrial assembly. 
Source code and demonstration videos are available at \mbox{\url{https://share-rl.github.io/share-rl.io/}}
\end{abstract}

\section{Introduction}
\begingroup
\footnotetext[1]{This work was supported by the Fraunhofer Internal Programs under Grant No. SME 40-09551.}
\endgroup
\setcounter{footnote}{1}
\footnotetext[1]{CITEC, Faculty of Technology, Bielefeld University, Germany.}
\footnotetext[2]{Fraunhofer IOSB-INA, Lemgo, Germany.}
\setcounter{footnote}{2}

High-mix low-volume (HMLV) industrial assembly, which is ubiquitous in small and medium-sized enterprises (SMEs), remains one of the most difficult challenges for autonomous robots~\cite{Peterson2020}. These settings demand near-zero failure rates, high precision across diverse product variants, and strict compliance with safety regulations, all while operating under inherent environmental uncertainty~\cite{Elguea-Aguinaco.etal2023,Braun.Wrede2023}. The challenge is exacerbated by the fact that SMEs typically lack automation experts, yet depend on production flexibility to remain competitive. Consequently, such tasks are still carried out almost exclusively by human operators. Growing labor shortages further intensify the need for robots that can adapt to changing assembly requirements, but today’s industrial systems remain notoriously inflexible.

\begin{figure}[h]
    \centering
    \includegraphics[width=\linewidth]{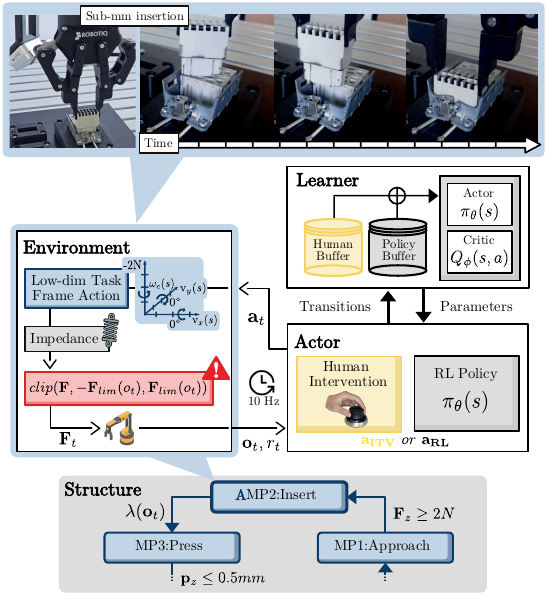}
    \caption{\textbf{Structure, interaction and safety.} SHaRe-RL is a hybrid approach that combines a hand-designed primitive sequence with interactive learning in selected primitives and adaptive per-axis force limits for bounded contact forces. 
    }
    \label{fig:visual_abstract}
    \vspace{-10pt}
\end{figure}

Reinforcement learning (RL) promises adaptable robots that can handle uncertainty and even outperform hand-designed controllers or teleoperation~\cite{Rajeswaran.etal2018}. However, deploying RL in real-world industrial settings remains difficult due to sample inefficiency, under- or overspecified reward functions, and the risks of unsafe exploration in contact-rich interactions~\cite{Vecerik.etal2018, Luo.etal2018}.

In order to close this gap, prior work has emphasized the importance of injecting domain knowledge into RL~\cite{Braun.Wrede2020}. Notably, two powerful forms of prior knowledge commonly exist in industrial settings: process structure, captured by production experts during engineering time through established formalisms~\cite{Finkemeyer.etal2005}, and operator expertise, provided through demonstrations and online corrections~\cite{Beierling.etal2024}. Each of these has proven effective in isolation~\cite{Braun.Wrede2023,Luo.etal2024}, but they have not been combined with modern, sample-efficient RL under real-world safety constraints.

We present SHaRe-RL, a Structured, Human-in-the-loop, and Real-world RL framework that integrates these sources of prior knowledge with a simple but effective compliance mechanism. Fig.~\ref{fig:visual_abstract} illustrates how SHaRe-RL embeds sample-efficient, interactive RL (yellow) within established task abstractions such as manipulation primitives (blue), while ensuring safety through adaptive force limits (red). SHaRe-RL scales to complex, vision-based assembly by guiding exploration to meaningful subspaces and focusing learning where it is most beneficial. Through demonstrations and online corrections, SHaRe-RL exploits operator expertise to accelerate exploration, while RL itself is able to improve beyond suboptimal human behavior. This hybrid approach preserves interpretability, supports effective human-robot collaboration, and enables faster, safer, and more economically viable training, moving RL a step closer to broad industrial acceptance. Our contributions are as follows:

\begin{itemize}
    \item We introduce SHaRe-RL as the first system to combine manipulation primitive nets with human-in-the-loop, off-policy reinforcement learning.
    \item We formulate an adaptive per-axis compliance mechanism that provably bounds interaction forces during exploration while preserving free-space dynamics.
    \item We demonstrate empirically that SHaRe-RL achieves robust performance on a sparse-reward industrial insertion task with sub-millimeter tolerances within three hours of real-world interaction, beating the state-of-the-art and even surpassing the cycle time of a skilled human demonstrator.
    \item We show generalization to previously unseen connector variants and thoroughly validate SHaRe-RL through ablations and analysis.
    \item We implement SHaRe-RL as an extension of the open-source LeRobot framework~\cite{Cadene.etal2024}, making it compatible with a wide range of low-cost manipulators.
\end{itemize}

\section{Related Work}
Contact-rich assembly is difficult to solve end-to-end with reinforcement learning, as large state–action spaces, sample efficiency, and strict safety constraints hinder direct trial-and-error learning~\cite{Elguea-Aguinaco.etal2023}. Robotics has therefore long relied on abstractions to manage complexity, such as manipulation primitives (MPs) and the task frame formalism~\cite{Mason1981,Kroger.etal2004,Bargmann.etal2024}. These abstractions can be composed into higher-level structures called Manipulation Primitive Nets (MP-Nets)~\cite{Finkemeyer.etal2005}. Early formulations that combine abstractions with learning-based methods treat primitives as discrete choices with small parameter sets to be optimized~\cite{Masson.etal2015,Dalal.etal2021}. Their process-level coordination can also be learned via hierarchical RL ~\cite{Vuong.Pham2023}. More recent approaches explore neural-network based primitives, first in isolation for compliant skills such as cutting~\cite{Padalkar.etal2020}, and later within assembly sequences~\cite{Braun.Wrede2023}. While these approaches demonstrate that task structure can drastically reduce exploration complexity, they typically assume on-policy updates, low-dimensional state features and carefully engineered reward functions. This limits their applicability to realistic industrial tasks with high-dimensional input and sparse rewards.

In parallel, advances in sample-efficient RL have shown that offline demonstrations and online interventions can drastically accelerate learning~\cite{Rajeswaran.etal2018, Luo.etal2018,Vecerik.etal2018}. Off-policy RL with demonstrations~\cite{Hu.etal2024,Rajeswaran.etal2018,Ball.etal2023}, model-based approaches~\cite{Nagabandi.etal2019,Wu.etal2022}, interactive or residual imitation learning~\cite{Ankile.etal2024,Ross.etal2011,Kelly.etal2019a}, and distributed frameworks like SERL~\cite{Luo.etal2024c} and its human-in-the-loop extension HIL-SERL~\cite{Luo.etal2024} demonstrate that human input can make sparse-reward, real-world tasks tractable. However, these approaches are typically trained and evaluated as isolated skills rather than as components of larger sequences and rarely exploit the forms of prior knowledge that are widely available in practice. Our contribution is to embed the same class of sample-efficient, human-in-the-loop RL into structured assembly processes. 

Traditional control approaches often achieve robustness but remain conservative and depend heavily on engineering expertise~\cite{Hartmann.etal2025}, whereas learning-based controllers promise greater adaptability but face challenges in safety and interpretability. Here, classical optimization-based safety layers, such as control barrier functions (CBFs), quadratic program (QP) filters, and model predictive control (MPC) shields, can enforce constraints during execution, but trade off model accuracy against computational overhead~\cite{Meng.etal2023,Dalal.etal2018,Wang.etal2024c}. Recent work instead learns variable-compliance schemes that adapt robot stiffness through RL, but lower stiffness alone cannot prevent unbounded contact forces when the policy issues large set-points, requiring careful tuning to avoid~\cite{Tahmaz.etal2025,Beltran-Hernandez.etal2020,Martin-Martin.etal2019}. In contrast, deterministic schemes that directly expose physically meaningful limits offer interpretability and ease of deployment. Our method contributes to this line by introducing per-axis adaptive force limits, which provably bound contact forces while preserving free-space dynamics, enabling RL to train safely in contact-rich settings.

\section{Method}
SHaRe-RL combines three complementary sources of domain knowledge into one asynchronous learning pipeline. In this section we first show how human demonstrations and real‑time operator interventions are incorporated into an off‑policy RL backbone; we then explain how task–frame manipulation primitives focus exploration on the phases that matter most; and we conclude with a simple yet provably safe compliance layer that bounds interaction forces.

\subsection{Operator Guidance: Human-in-the-Loop RL}
Robotic reinforcement learning tasks can be formulated as a finite-horizon Markov Decision Process (MDP) where the objective is to learn a policy which maximizes the expected sum of discounted future rewards~\cite{Sutton.Barto2018}. We adopt a distributed human-in-the-loop RL training loop similar to HIL-SERL~\cite{Luo.etal2024}, which has proven effective for real-world RL with sparse rewards. 

Before training begins, a skilled operator collects a set of teleoperated trajectories with a six–DoF SpaceMouse, which are stored in a demonstration buffer. During training, the operator can press a button to take control of the robot at any time. While the button is held, the policy action is replaced by the human SpaceMouse action. This biases data collection toward successful and safe behaviors and helps the policy escape poor local optima~\cite{Luo.etal2024}. Beyond corrections, the operator can also deliberately induce rare but relevant failure cases, forcing the policy to learn robust recovery behavior. In practice, operators intervene frequently during the early stages of training and reduce their involvement as performance improves, with faster convergence when corrections are short and specific while otherwise letting the policy explore on its own.

At the system level, an actor process executes the current policy on the robot, generates experience, and streams it to the learner. The learner process samples mini-batches equally from the demonstration buffer and an on-policy buffer to perform gradient updates on the RLPD loss~\cite{Ball.etal2023}.  RLPD is a regularized variant of SAC~\cite{Haarnoja.etal2018} designed for sample efficiency and integration of prior data without offline pretraining. Updated network weights are streamed back to the actor at fixed time intervals.
\vspace{0.5em}

\subsection{Task-Level Priors: TFF and MP-Nets}
Human-in-the-loop corrections alone are not sufficient to enable long-horizon assembly, as the search space quickly becomes intractable without additional structure. A widely used strategy to add such structure is to decompose complex processes into manipulation primitives (MPs), each representing a fundamental capability of the manipulator~\cite{Suarez-Ruiz.Pham2015}. Finkemeyer et al.~\cite{Finkemeyer.etal2005} formalize MPs as the following tuple
\[
    \text{MP} := \{ \text{HM}, \tau, \lambda \},
\]
where $\text{HM}$ describes a hybrid force- or position control policy for each translational and rotational direction relative to a task frame, $\tau$ is a tool command and $\lambda$ is a boolean predicate on sensor values that defines a stop condition~\cite{Mason1981,Kroger.etal2004}. Importantly, the type of controller can differ in all directions. Manipulation Primitive Nets (MP-Nets) extend this abstraction by composing MPs into state machines, where transitions are triggered by stop conditions~\cite{Finkemeyer.etal2005}.

Classical MPs are executed with fixed control setpoints. Extending this notion, adaptive MPs (AMPs) replace selected setpoints by policy-controlled variables~\cite{Braun.Wrede2023}. If $n$ axes are controlled by the policy, the corresponding action space is $n$-dimensional. The operator’s SpaceMouse is likewise constrained to these $n$ axes, ensuring that both demonstrations and interventions remain aligned with the degrees of freedom available to the policy. This formulation focuses exploration on task-relevant directions and high-reward regions of the state-space. For example, in peg-in-hole insertion, the coarse approach is executed with fixed setpoints, while learning is reserved to handle uncertainty during fine alignment. 

SHaRe-RL uses end-to-end demonstrations as a complementary source of prior knowledge to improve the sample efficiency of AMPs. Instead of training individual MPs in isolation with their own reset strategy, both demonstrations and online rollouts traverse the entire MP-Net from initial to terminal state. During data collection the operator teleoperates adaptive primitives with the SpaceMouse, while non-adaptive primitives run with fixed set-points; all predefined stop conditions are active.  When a stop event is difficult to encode analytically, the operator can signal the end of a primitive with a button press. These labels are used to train a lightweight binary classifier that later triggers the same transition autonomously, which is common practice in real-world RL~\cite{Singh.etal2019}.

This design has three advantages. First, MP-Nets can explicitly encode reset strategies, such that the same graph that executes a task also returns the system to a well-defined start state. In our experiments, the MP-Net includes an explicit reset primitive from the terminal back to the start state, enabling fully automatic, reset-free rollouts without manual intervention~\cite{Gupta.etal2021}. Second, our software stack supports jointly optimizing multiple AMPs within one MP-Net, though we focus on a single AMP in this work and leave multi-primitive joint optimization to future work. Third, the initial state distribution of each AMP is correctly induced by its predecessors. Specifically, uncertainty introduced by upstream actions, such as variability in a grasp, must be handled robustly by downstream primitives. Training with “atomic” reset strategies would alter these distributions and yield policies that generalize poorly at execution time. Training with full MP-Net rollouts instead exposes policies to the exact state distributions they will encounter, improving robustness during deployment.  

\vspace{0.5em}
\subsection{Adaptive Safety Limits}
\begin{figure}[t]
    \centering
    \includegraphics[width=\linewidth]{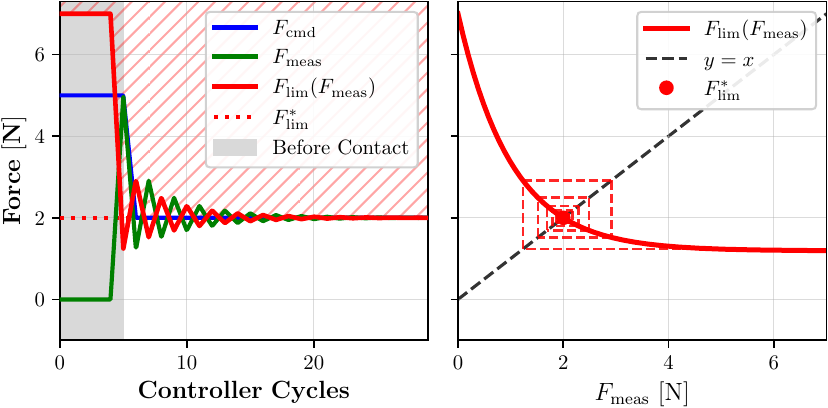}
    \caption{\textbf{Adaptive force limits under ideal conditions} Left: Time response of a stable adaptive limit. Right: Phase-plane cobweb plot illustrating the recurrence in Eq.\ref{eq:recurrence}.}
    \label{fig:forces_ideal}
\end{figure}

Exploration in contact-rich assembly can easily produce large, destructive contact forces. Beyond the risk of hardware damage, such events degrade sample efficiency by aborting episodes early and biasing trajectory distributions. To address this, we introduce adaptive per-axis force limits as a deterministic safety layer for compliant control. This mechanism enforces physically meaningful force bounds under contact without compromising responsiveness in free-space.

Let $F_{\mathrm{cmd},t}$ denote the raw output of the high-level controller or policy, and $F_{\mathrm{meas},t}$ the measured contact force at time step $t$. Given a static hardware bound $F_{\max}>0$, we define a time-varying adaptive $F_{\mathrm{lim},t}$
\begin{equation}
   F_{\mathrm{lim},t}(F_{\mathrm{meas},t}) = \alpha(|F_{\mathrm{meas},t}|)\,F_{\max} > F_{\mathrm{cmd},t},
  \label{eq:lim}
\end{equation}
\noindent where $\alpha:\mathbb{R}_{\ge 0}\!\to\!(0,1]$ is a monotone decreasing map with $\alpha(0)=1$. This limit is large during free-space motion, and contracts once contact is detected (\( F_{\mathrm{meas},t} > 0 \)). \newpage
\noindent Assuming rigid-body contact ($F_{\text{meas},t+1}=F_{\text{cmd},t}$), the closed loop evolves as
\begin{equation}
  F_{\mathrm{meas},t+1} = F_{\mathrm{lim},t} = \alpha(|F_{\mathrm{meas},t}|)\,F_{\max}
  \label{eq:recurrence}
\end{equation}
\noindent The long-term behavior of this recurrence is governed by its fixed point \( F^{*}_{\mathrm{lim}} \), which corresponds to the steady-state force established under contact. Convergence to \( F^{*}_{\mathrm{lim}} \) is guaranteed whenever the recurrence is locally stable, i.e., it satisfies

\begin{equation}
  \left| \alpha'\left(F^{*}_{\mathrm{lim}}\right) \right| F_{\max} < 1
  \label{eq:stability}
\end{equation}
A practical choice for \( \alpha \) is an exponential decay with minimal scaling \( \alpha_{\min} \):
\begin{equation}
  \alpha(F) = \alpha_{\min} + (1-\alpha_{\min}) \exp(-F/\theta).
  \label{eq:exp}
\end{equation}
\noindent This form is smooth, monotone, and admits a unique fixed point, which can be set deterministically by solving for the corresponding decay parameter \( \theta^* \) given \( F_{\max} \) and a desired \( F^{*}_{\mathrm{lim}} \). Stability then reduces to a simple inequality on \( \alpha_{\min} \), which is conceptually similar to contraction-based verification of dynamical systems~\cite{Neumann.Steil2015}. Fig. \ref{fig:forces_ideal} illustrates the idealized time and phase-plane response of a stable adaptive force limit. Large commands are allowed in free space, but once $F_{\text{meas}}$ exceeds zero the limit shrinks and both commanded and measured forces settle at \( F^{*}_{\mathrm{lim}} \) without manual tuning.

\begin{figure}[t]
    \centering
    \includegraphics[width=\linewidth]{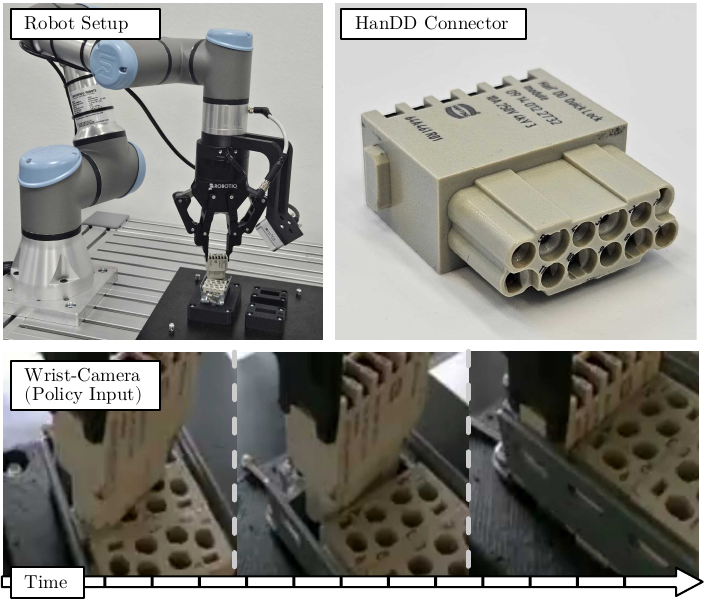}
    \caption{\textbf{Experimental setup and task.} Top: UR3e robot and Harting HanDD industrial connector used for evaluation. Bottom: Cropped policy input over time.}
    \label{fig:setup}
\end{figure}

\section{Connector Assembly Task}
\begin{figure}[t]
    \includegraphics[width=\linewidth]{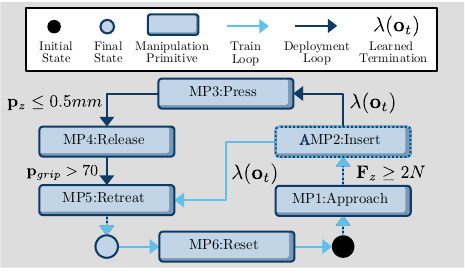}

    \vspace{3pt}
    
    \resizebox{\linewidth}{!}{
    \begin{tabular}{c|cccccc}
    \toprule
    \textbf{ID} & \textbf{x} & \textbf{y} & \textbf{z} & \textbf{a} & \textbf{b} & \textbf{c} \\
    \midrule
    MP1 & \SI{0}{\milli\metre\per\second} & \SI{0}{\milli\metre\per\second} & \SI{-10}{\milli\metre\per\second} & \SI{0}{\degree} & \SI{0}{\degree} & \SI{0}{\degree\per\second} \\
    AMP2 & $v_x(\mathbf{o}_t)$~\si{\milli\metre\per\second}
        & $v_y(\mathbf{o}_t)$~\si{\milli\metre\per\second}
        & \SI{2}{\newton} & \SI{0}{\degree} & \SI{0}{\degree} 
        & $\omega_c(\mathbf{o}_t)$~\si{\degree\per\second} \\
    MP3 & \SI{0}{\milli\metre\per\second} & \SI{0}{\milli\metre\per\second} & \SI{1}{\milli\metre\per\second} 
        & \SI{0}{\degree} & \SI{0}{\degree} & \SI{0}{\degree\per\second} \\
    MP4 & \SI{0}{\milli\metre\per\second} & \SI{0}{\milli\metre\per\second} & \SI{0}{\milli\metre} 
        & \SI{0}{\degree} & \SI{0}{\degree} & \SI{0}{\degree\per\second} \\
    MP5 & \SI{0}{\milli\metre\per\second} & \SI{0}{\milli\metre\per\second} & \SI{25}{\milli\metre} 
        & \SI{0}{\degree} & \SI{0}{\degree} & \SI{0}{\degree\per\second} \\
    MP6 & $\mathcal{N}(20, 5)$ \SI{}{\milli\metre} & $\mathcal{N}(0, 5)$ \SI{}{\milli\metre} & \SI{25}{\milli\metre} 
        & $\mathcal{N}(0, 10)$\SI{}{\degree} & $\mathcal{N}(0, 10)$\SI{}{\degree} & $\mathcal{N}(0, 10)$\SI{}{\degree} \\
    
    \end{tabular}
    }
    \caption{\textbf{MP-Net for the HanDD connector insertion.} Light-blue and dark-blue arrows indicate the train and deployment loop respectively. The learned termination condition $\lambda(\mathbf{o}_t)$ and the randomized reset MP6 close the loop for repeated trials. The table lists the per-axis controller setpoints for each primitive.}
    \label{fig:mp_net}
    \vspace{-1em}
\end{figure}

We evaluate our approach on the insertion of industrial Harting Han-Modular connectors, a representative contact-rich assembly task with tolerances of only 0.2–0.4 mm. The task resembles peg-in-hole insertion but involves non-cylindrical geometry and a final snap-in stage. While the learning problem itself is concentrated in the fine insertion phase, reliable real-world training benefits from structuring the overall process into a small number of manipulation primitives. Scripted primitives handle the repeatable parts of the process (approach, coarse alignment and post-contact actions), while learning is restricted to the uncertain insertion where it is most beneficial. Fig.~\ref{fig:setup} illustrates the setup, and Fig.~\ref{fig:mp_net} summarizes the MP-Net used in our experiments, including an explicit reset primitive (MP6) that returns the robot from terminal back to the start state and enables fully automatic rollouts.

\subsection{Robot and Controller}
Experiments are conducted on the UR3e depicted in Fig.~\ref{fig:setup}, which is equipped with an integrated six-axis force/torque sensor at the wrist, an ego-centric Intel RealSense D405 RGB-D camera, and custom 3D-printed gripper fingers. The control hierarchy consists of three layers. At the top, desired force or velocity setpoints in task frame axes are updated at 10~Hz. These targets are processed by a mid-level Cartesian impedance loop running at 500~Hz, which translates them into wrench commands for the robot’s proprietary force controller. This formulation enables us to apply per-axis force limits directly to the commanded wrench before it reaches the hardware. In free space, the maximum limits are set to \SI{30}{\newton} for translation and \SI{10}{\newton\meter} for rotation, while under contact the adaptive scaling settles at \SI{5}{\newton} and \SI{0.5}{\newton\meter}, respectively, ensuring safety during exploration.

\subsection{MDP Formulation}
The connector insertion is modeled as a finite-horizon MDP. To provide a clearer formalization, we decompose this into the observation and action space, reset mechanism, and reward function. 

\textbf{Observation Space:} The observation at time $t$ is defined as 
\begin{equation}
\mathbf{o}_t = \big(I_{t}, \mathbf{x}_{t-1:t}, \mathbf{F}_{t-1:t}\big),
\end{equation}
\noindent where $I_t \in \mathbb{R}^{128\times128\times3}$ is the wrist-mounted RGB image, $\mathbf{x}_{t-1:t} \in \mathbb{R}^{2\times12}$ are stacked end-effector poses and velocities, and $\mathbf{F}_{t-1:t} \in \mathbb{R}^{2\times6}$ are stacked wrenches. Frame stacking compensates for partial observability and non-Markovian effects introduced by the compliant controller. To prevent overfitting to a fixed geometry, absolute positions along learned axes are excluded from the state input. Images are cropped and downsampled to a local view, augmented with random shifts, and encoded with a frozen, pretrained ResNet10~\cite{Hansen.Wang2021,He.etal2015}. The resulting embeddings are concatenated with proprioceptive and wrench features before being processed by the policy and critic networks. Both networks are implemented as two-layer multilayer perceptrons (MLPs) with 256 hidden units per layer and SiLU activations~\cite{Elfwing.etal2017}. The policy outputs the parameters of a diagonal Gaussian, which is squashed with \texttt{tanh} to enforce action limits, following the standard practice from Soft Actor-Critic~\cite{Haarnoja.etal2018}. Unless stated otherwise, all methods in Table~\ref{tab:main} receive the observation $\mathbf{o}_t$ and use the same network architecture, preprocessing, and augmentation pipeline.

\textbf{Action Space:} The action space depends heavily on the structural priors provided to the agent. For our structured variants, the MDP is defined solely over the insertion primitive. Consequently, the action space is restricted to Cartesian velocity setpoints for the $x$-, $y$-, and $c$-axes. In contrast, the unstructured baselines must solve the insertion end-to-end without task-level priors, requiring them to act in the full six-DoF Cartesian space.

\textbf{Reset:} All methods share the same scripted reset mechanism and episode horizon. Compared to prior work~\cite{Braun.Wrede2023}, we expose an additional rotational degree of freedom around the connector’s $c$-axis and increase initial pose uncertainty. At every episode start, Gaussian noise is applied independently to each axis: ($\sigma_{x,y}=5$\,mm) in translation and ($\sigma=10^\circ$) in each rotation axis. To simulate systematic calibration errors, the initial distribution is offset by 2\,cm along the x-axis. As a result, the policy must traverse a longer distance before fine alignment, increasing the maximum episode length from 3\,s to 9\,s (90 steps).

\textbf{Reward Function:} Finally, the learning objective is defined by either a dense shaping reward based on vertical distance to the target, or a sparse binary success classifier trained offline:
\[
r_t^{\text{dense}} = -\,\|z_t - z_{\text{goal}}\|_2,
\qquad
r_t^{\text{sparse}} = \lambda(\mathbf{o}_t),
\]
\noindent The classifier \( \lambda(\mathbf{o}_t) \) uses the same network architecture as the policy and is trained offline before RL on a mix of 20 positive and 20 negative demonstrations. Since many industrial success signals are inherently sparse, and sparse classifiers tend to generalize better under uncertainty~\cite{Schoettler.etal2019}, we evaluate on \( r_t^{\text{sparse}} \) for our main experiments and use \( r_t^{\text{dense}} \) for ablations.

\begin{figure}[t]
    \centering
    \includegraphics[width=\linewidth]{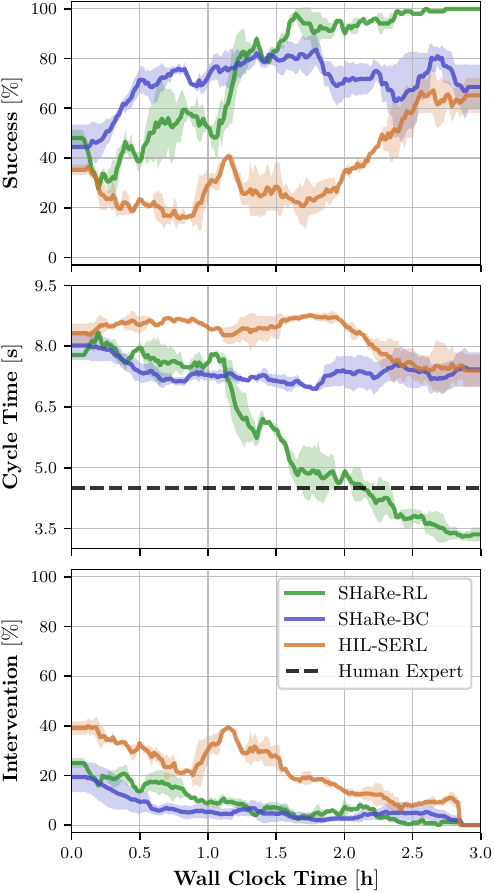}
    \caption{\textbf{Main comparison of learning approaches.} Success rate, cycle time, and intervention rate vs wall-clock time (mean ± s.e.m., 2 seeds).}
    \label{fig:main}
\end{figure}

\vspace{10px}
\section{Experimental Results}
A key premise of our method is that task-level motion primitives drastically reduce the size of the effective state-action space, and thereby lower the overall complexity of the learning problem. We evaluate this hypothesis by comparing structured interactive RL (SHaRe-RL) to (i) an unstructured human-in-the-loop RL baseline (HIL-SERL) that does not use MPs, and (ii) a structured interactive behavior cloning baseline (SHaRe-BC) similar to HG-Dagger~\cite{Kelly.etal2019a}. In addition, we include pure RL baselines and structured ablations to isolate the role of demonstrations, interventions, and vision. We compare all methods over two training runs using a fixed wall-clock budget of three hours on a single NVIDIA RTX~3060~Ti workstation. Each variant is initialized with 20 teleoperated demonstrations and evaluated at the end of training using 20 autonomous rollouts without interventions. During training, we track success rate, cycle time, and intervention rate using a rolling average. The policy executes at 10 Hz; cycle time is measured from the beginning of a rollout after reset until the success condition is met or the episode times out. For structured methods, this definition includes the approach primitive and is therefore comparable across structured and unstructured methods.

\begin{table}[t]
\centering
\scriptsize 
\setlength{\tabcolsep}{3pt} 
\begin{tabular}{l|cc|ccc}
\toprule
Method & Success [\%] & Time [s] & Structure & Demos & Interventions \\
\midrule
SAC$_{\text{dense}}$ & 5 & 8.7 & \xmark & \xmark & \xmark \\
HIL-SERL & 65 & 7.4 & \xmark & \cmark & \cmark \\
SHaRe-BC & 80 & 7.4 & \cmark & \cmark & \cmark \\
\textbf{SHaRe-RL} & \textbf{100} & \textbf{3.5} & \cmark & \cmark & \cmark \\
\midrule
SHaRe-RL$_{\text{PPO}}$ & 5 & 8.8 & \cmark & \xmark & \xmark \\
SHaRe-RL$_{\text{no-demos}}$ & 35 & 7.9 & \cmark & \xmark & \xmark \\
SHaRe-RL$_{\text{no-interventions}}$ & 45 & 7.5 & \cmark & \cmark & \xmark \\
SHaRe-RL$_{\text{no-vision}}$ & 10 & 8.8 & \cmark & \cmark & \cmark \\
\midrule
Human Expert & 100 & 4.5 & -- & -- & -- \\
Random Policy & 5 & 8.9 & -- & -- & -- \\
\bottomrule
\end{tabular}
\caption{\textbf{Main comparison across methods.} SHaRe-RL variants use the same off-policy RL backbone but differ in which sources of prior knowledge are used. HIL-SERL represents human-in-the-loop RL without task-level priors, while SHaRe-BC represents structured interactive imitation without reinforcement learning.}
\label{tab:main}
\end{table}

\subsection{Main Results}

Fig.~\ref{fig:main} and Table~\ref{tab:main} summarize the main comparison. SHaRe-RL achieves the best overall performance, reaching 100\% success with a 3.5\,s cycle time, and thus outperforming both the human expert (4.5\,s at 100\%) and all learning baselines. Importantly, SHaRe-RL also improves substantially over the unstructured human-in-the-loop RL baseline HIL-SERL, which reaches 65\% success at 7.4\,s. This gap directly quantifies the practical benefit of adding structure to contact-rich real-world RL. SHaRe-BC quickly reaches strong performance (80\% at 7.4\,s), but plateaus near the quality of its demonstrations, whereas SHaRe-RL continues to improve throughout the full training budget. This reflects the core strengths of RL over BC: human demonstrations can be suboptimal due to pauses, conservative behavior or operator tendencies. For example, we observed that humans typically align each axis sequentially when using a spacemouse, resulting in longer trajectories. BC performance is bounded by the quality of the demonstrations, while RL can use dynamic programming to learn from both successes and failures. This allows SHaRe-RL to self-correct, discover behaviors that are hard to demonstrate, such as simultaneous multi-axis alignment with optimal dynamics, and ultimately exceed human performance. This is consistent with theoretical results~\cite{Luo.etal2024b}.

We also evaluate intervention rates during training to measure the shift toward autonomous execution. Specifically, we compute the ratio of intervened timesteps to total timesteps per episode, averaged over a 20-episode window. As shown in Fig.~\ref{fig:main}, SHaRe-RL's reliance on human corrections steadily decreases and reaches zero by the end of training. Qualitatively, we observe a shift from long, sparse interventions early in training to brief, targeted corrections as the policy matures.

\subsection{Ablation Study}
The ablations in Table~\ref{tab:main} disentangle the contributions of demonstrations, interventions, and vision. Removing online corrections (SHaRe-RL$_{\text{no-interventions}}$) reduces performance to 45\% success, while further removing offline demonstrations (SHaRe-RL$_{\text{no-demos}}$) only achieves 35\%, confirming that both sources of human guidance are important within the three-hour budget. The on-policy variant (SHaRe-RL$_{\text{PPO}}$)~\cite{Schulman.etal2017} reaches just 5\% success after the full training budget, highlighting its inability to handle sparse rewards and vision input. Removing vision (SHaRe-RL$_{\text{no-vision}}$) collapses performance to 10\%, indicating that image feedback is critical for resolving geometric ambiguity, in particular around the connector’s $c$-axis, which cannot be inferred reliably from force and velocity alone under the pose uncertainties considered.

\begin{figure}[b]
    \centering
    \includegraphics[width=\linewidth]{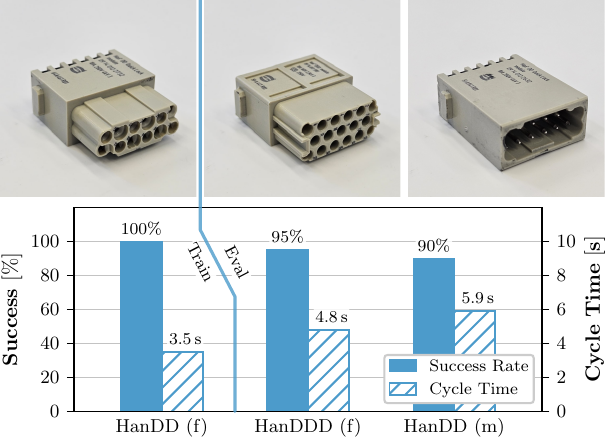}
    \caption{\textbf{Generalization across novel connectors.} A policy trained on HanDD (f) is evaluated without finetuning on two previously unseen connector variants. We report success rate and cycle time over 20 autonomous rollouts per connector.}
    \label{fig:generalization}
\end{figure}

\begin{figure}[t]
    \centering
    \includegraphics[width=\linewidth]{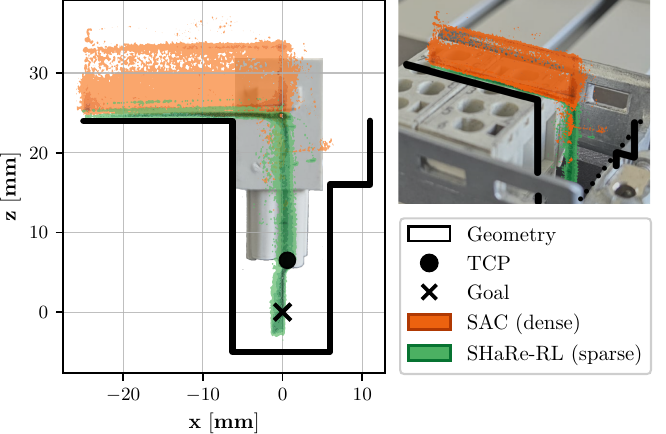}
    \caption{\textbf{State-space occupancy during training.} Fully autonomous rollouts projected onto a 2D $x–z$ slice of the workspace. SAC (dense), trained end-to-end without task structure, explores broadly but rarely enters the narrow funnel that leads to successful insertion. SHaRe-RL (sparse) concentrates occupancy in goal-relevant regions.}
    \label{fig:occupancy}
\end{figure}

To better understand the role of task and operator priors, we additionally visualize state-space occupancy during training in Fig.~\ref{fig:occupancy}. We compare strictly autonomous rollouts of SHaRe-RL to an unstructured dense-reward baseline (SAC$_{\text{dense}}$) that is trained end-to-end without primitives, demonstrations, or interventions. Although the dense reward provides a continuous learning signal along the insertion direction, uncertainty in orientation and contact geometry renders this signal largely uninformative for alignment. As a result, SAC$_{\text{dense}}$ explores a much larger region but remains centered above the opening and rarely enters the narrow funnel required for insertion. The policy fails to align the connector reliably, as the reward landscape is largely flat along rotational axes. In contrast, SHaRe-RL concentrates occupancy in the successful region even under sparse rewards, indicating that well-chosen priors can be more effective than hand-crafted shaping rewards for guiding exploration in contact-rich settings.

\subsection{Zero-Shot Generalization}
We further evaluate the zero-shot transfer capabilities of the learned policy by deploying it, without any finetuning, on two previously unseen connector variants (Fig.~\ref{fig:generalization}). Although the average cycle time increases as the geometry and contact dynamics change, the success rate remains robust. This demonstrates that SHaRe-RL learns a generalized insertion strategy rather than memorizing a specific spatial configuration. Prior work has shown that transferring policies across connector geometries is difficult when relying solely on force feedback~\cite{Braun.Wrede2023}, supporting the necessity of image-based observations for robust assembly.

\subsection{Adaptive Force Limits}
For a task-level comparison, we compare our adaptive limits to the HIL-SERL-style limits that bound the impedance reference error and thus bound the maximum wrench at all times~\cite{Luo.etal2024}. During demonstration collection, adaptive limits achieve 100\% success with 4.5\,s average cycle time, compared to 90\% success and 5.2\,s with HIL-SERL limits, where failures are timeouts at the episode horizon. Limits are applied independently per axis; in our insertion setup contact is dominated by the $z$ direction, and we did not observe adverse effects from cross-axis coupling.

\begin{figure}[t]
    \centering
    \includegraphics[width=\linewidth]{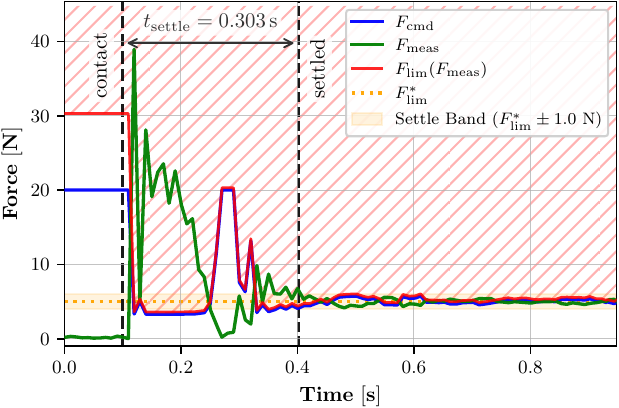}
    \vspace{-1.6em}
    \caption{\textbf{Adaptive force limits on the real system in a high-force collision test.}
    We command 20\,N in $z$ against a rigid stop with $F_{\max}=30$\,N and $F^*_{lim}=5$\,N. Prior to contact, large commands are permitted; upon contact, the limit contracts and the measured force converges to $F^*_{lim}$.}
    \label{fig:forces}
\end{figure}

We further evaluate the limits in a high-force collision test by commanding 20\,N in the $z$-direction against a rigid stop. Performance is averaged over 20 independent trials (Fig.~\ref{fig:forces}). 
Despite measurement noise and signal smoothing in the hierarchical controller, the measured force reliably settles within a $\pm$1.0 N band around the 5.0 N target in an average of 0.303 s. Once settled, the controller demonstrates consistent stability, maintaining a steady-state mean error of 0.004 N with a standard deviation of 0.37 N. This confirms that the adaptive limits effectively mitigate impact energy and bound contact forces safely during exploration.

\section{Conclusion}
This work introduced SHaRe-RL, a modular reinforcement learning framework showing how structured priors and interactive learning together overcome the limitations of both hand-engineered control and unstructured trial-and-error learning. Our experiments on industrial connector insertion show that this formulation improves over an unstructured human-in-the-loop RL baseline and can generalize to previously unseen connector variants. This suggests that, when guided by domain expertise already present in SMEs, RL can be both sample-efficient and safe enough for practical use, which substantially lowers the barrier to deploying learning in contact-rich industrial assembly.

A key challenge that remains is the amount of human supervision required for complex tasks. This motivates future work on more autonomous methods, such as model-based exploration~\cite{Watters.etal2019}, or self-supervised reward signals from images~\cite{Ma.etal2023}. Our method is also orthogonal to other means of reducing real-world training time, such as offline RL pretraining~\cite{Nair.etal2023}, sim-to-real transfer~\cite{Bergmann.etal2024}, and the integration of large datasets and pretrained models~\cite{Collaboration.etal2024}.

Looking ahead, while SHaRe-RL performs well within a connector family, HMLV deployment ultimately requires broader generalization across product variants and operating conditions. A natural next step is to leverage pretrained foundation models. Recent vision-language-action (VLA) systems~\cite{Intelligence.etal2025a} suggest that combining large-scale pretraining, offline RL objectives, and human-in-the-loop rollouts can improve over dataset performance, yet whether such recipes transfer to specialized industrial assembly, where tasks are strongly out-of-distribution and in-domain data is limited, remains open. Identifying a robust RL fine-tuning recipe in this regime is therefore a central next step for scaling SHaRe-RL beyond narrowly scoped skills. Second, we aim to reduce the burden of manually specifying process structure. While MP-Nets provide sample efficiency and interpretability, designing them still requires manual engineering effort. An important direction is to learn an initial, human-interpretable primitive graph from end-to-end demonstrations and iteratively refine it with human feedback, implementing a form of co-construction~\cite{Vollmer.etal2014} that spans from low-level control to high-level task programming.

\bibliographystyle{IEEEtran} 
\bibliography{sharerl,bstctl}

\end{document}